\documentclass{article}

\usepackage[preprint]{neurips_2025}

\usepackage[utf8]{inputenc} 
\usepackage[T1]{fontenc}    
\usepackage{hyperref}       
\usepackage{url}            
\usepackage{booktabs}       
\usepackage{amsfonts}       
\usepackage{nicefrac}       
\usepackage{microtype}      
\usepackage{xcolor}         

\usepackage{float}

\usepackage{xspace}

\usepackage{subcaption}
\usepackage{graphicx}
\usepackage{svg}
\usepackage{wrapfig}

\usepackage{amsmath}

\usepackage[capitalise]{cleveref}

\newcommand{\diagramscale}{0.8}
\newcommand{\plotscale}{0.42}
\newcommand{\fphantom}{\vphantom{\left[\frac{1}{1}\right]}}
\svgsetup{inkscapepath=svgsubdir}  

\newcommand{\twinturbo}{TwinTURBO\xspace}

\newcommand{\twinturbotitle}{\twinturbo: Semi-Supervised Fine-Tuning of Foundation Models via Mutual Information Decompositions for Downstream Task and Latent Spaces\xspace}

\newcommand{\EE}{\mathbb{E}}
\renewcommand{\L}{\mathcal{L}}

\newcommand{\kld}[2]{D_{\mathrm{KL}}\left(#1\,||\,#2\right)}
\newcommand{\jsd}[2]{D_{\mathrm{JS}}\left(#1\,||\,#2\right)}
\makeatletter
\def\E_#1{\EE_{#1}\@ifnextchar[{\Ebrac}{\relax}}
\def\Ebrac[#1]{\left[#1\right]}
\makeatother

\newcommand{\x}{\mathbf{x}}
\newcommand{\y}{\mathbf{y}}
\newcommand{\z}{\mathbf{z}}

\newcommand{\zt}{\tilde{\z}}

\newcommand{\yh}{\hat{\y}}

\newcommand{\X}{\mathbf{X}}
\newcommand{\Y}{\mathbf{Y}}
\newcommand{\Z}{\mathbf{Z}}

\setcitestyle{numbers}
\setcitestyle{square}
\setcitestyle{unsorted}

\begin{document}

\title{\twinturbotitle}

\author{%
  Guillaume Quétant\thanks{Corresponding authors: \texttt{guillaume.quetant@unige.ch}, \texttt{svolos@unige.ch}} \\
  Dept. of Computer Science \\
  University of Geneva \\
  Carouge, Switzerland \\
  \And
  Pavlo Molchanov \\
  NVIDIA \\
  \And
  Slava Voloshynovskiy$^*$ \\
  Dept. of Computer Science \\
  University of Geneva \\
  Carouge, Switzerland \\
}


\maketitle

\begin{abstract}
We present a semi-supervised fine-tuning framework for foundation models that utilises mutual information decomposition to address the challenges of training for a limited amount of labelled data.
Our approach derives two distinct lower bounds: i) for the downstream task space, such as classification, optimised using conditional and marginal cross-entropy alongside Kullback-Leibler divergence, and ii) for the latent space representation, regularised and aligned using a contrastive-like decomposition.
This fine-tuning strategy retains the pre-trained structure of the foundation model, modifying only a specialised projector module comprising a small transformer and a token aggregation technique.
Experiments on several datasets demonstrate significant improvements in classification tasks under extremely low-labelled conditions by effectively leveraging unlabelled data.

\end{abstract}

\section{Introduction}

Foundation models are large-scale neural networks pre-trained on diverse data to learn general-purpose representations that can be fine-tuned for specific downstream tasks.
This poses significant challenges, especially in the case of low-labelled data, a semi-supervised learning setting where only a small fraction of the data samples are labelled, while the majority remain unlabelled.
While foundation models are pre-trained on large datasets in a self-supervised manner, their deployment often requires fine-tuning on new datasets with limited labelled samples and potential distribution shifts.
Furthermore, the downstream tasks frequently differ from the pre-training objectives, complicating the adaptation process.
Existing semi-supervised approaches, such as pseudo-labelling, rely heavily on assumptions about data distributions or task-specific tuning, limiting their generalisability.
Addressing these challenges is essential to fully exploit the potential of foundation models and ensure their adaptability and scalability in diverse applications.
The main contributions of this study are:

\begin{itemize}
    \item \textbf{A new framework for foundation models fine-tuning:} We introduces a fine-tuning strategy based on mutual information decomposition.
    This framework guides the adaptation of foundation models to downstream tasks and latent space representations, providing a theoretically sound and efficient alternative to traditional supervised learning.
    
    \item \textbf{Novel theoretical insights:} The framework derives two distinct mutual information bounds: one for optimising downstream task performance and another for aligning latent space representations.
    These insights deepen our understanding of how information can be effectively utilised in semi-supervised learning.
    
    \item \textbf{Practical validation:} Extensive experiments on classification tasks under extremely low-labelled data conditions demonstrate significant performance improvements.
    These results highlight the effectiveness of focusing on theoretical decompositions over heuristic-based approaches, and reveal previously underexplored opportunities in semi-supervised fine-tuning.
\end{itemize}

\twinturbo builds on and extends existing work in semi-supervised learning and mutual information-based optimisation \cite{taran2020variationalibn,quetant2023turbo,drozdova2024semisupervised}.
Traditional approaches often emphasise pseudo-label generation or global fine-tuning strategies, which can be computationally expensive and less adaptable to diverse tasks.
Our formalism provides a lightweight and modular framework that combines pre-trained vision foundation models with a specialised adapter module.
This approach integrates with state-of-the-art methods such as InfoNCE-based contrastive learning \cite{oord2018infonce,chen2020simclr,caron2021dino,radford2021clip}, while addressing their limitations by providing a unified perspective on labelled and unlabelled semi-supervised fine-tuning.


\section{Related work}

\paragraph{Semi-Supervised Machine Learning.}

Semi-supervised learning methods aim to exploit large amounts of unlabelled data to improve performance when labelled data is scarce.
Traditional approaches include pseudo-labelling \cite{lee2013pseudo}, consistency regularisation \cite{sajjadi2016regularization, laine2017temporal}, and entropy minimisation \cite{grandvalet2005semi}. 
In the context of semi-supervised learning for foundation models, recent works explore the fine-tuning of pre-trained models with minimal labelled data while maximising the use of unlabelled samples.
Methods such as FixMatch \cite{sohn2020fixmatch} and FlexMatch \cite{zhang2021flexmatch} refine pseudo-labelling by enforcing consistency under data augmentations.
These approaches aim to minimise the reliance on large labelled datasets while maintaining strong generalisation capabilities.

\paragraph{Fine-Tuning of Foundation Models.}

Full-model fine-tuning of large-scale models has been shown to be effective, but computationally expensive.
To mitigate the computational cost of full-model fine-tuning, several works introduce adapter-based methods where additional small modules are trained while the pre-trained backbone is frozen.
LoRA \cite{hu2021lora}, VeRA \cite{kopiczko2023vera}, DoRA \cite{mao2024dora} and others introduce low-rank adaptation layers that allow efficient task-specific tuning without updating the entire model.
Other studies, such as Compacter \cite{mahabadi2021compacter} and Prompt Tuning \cite{lester2021power} have demonstrated that tuning a fraction of the parameters can achieve competitive results compared to full-model fine-tuning, especially in low-data regimes.
We also refer the reader to \cite{drozdova2024semisupervised} for a recent study on fine-tuning of vision foundation models.

\paragraph{Mutual Information for Machine Learning.}

Mutual information bounds play a crucial role in modern machine learning by formalising the relationship between observed data, latent representations and downstream task labels.
Classical approaches, such as the information bottleneck \cite{tishby2000information}, propose to minimise mutual information between inputs and latent variables while maximising mutual information between latent representations and outputs. 
More recent works decompose mutual information for machine learning applications.
The study presented in \cite{barber2004algorithm} introduces a variational lower bound on mutual information.
Other information-theoretic studies, such as the variational information bottleneck \cite{alemi2016deepvariational}, the exploration of several bounds \cite{poole2019variational,tschannen2019mutualinformation} and Deep InfoMax \cite{hjelm2018deepinfomax} apply mutual information maximisation to learn robust representations.
More recently, TURBO \cite{quetant2023turbo} also uses mutual information maximisation to give a meaningful interpretation of a wide family of deep learning models.
InfoNCE \cite{oord2018infonce} and its extensions provide tractable mutual information estimates via contrastive learning that is a common basis for modern unimodal self-supervised methods such as SimCLR \cite{chen2020simclr} and DINO \cite{caron2021dino} and multimodal ones such as CLIP \cite{radford2021clip}.

\section{\twinturbo}
\label{sec:twinturbo}

\twinturbo is a framework for guiding semi-supervised fine-tuning of foundation models using mutual information decompositions.
It adapts models to downstream tasks or latent spaces by leveraging data and statistical structures, bridging theoretical insights with practical applications, while connecting to and advancing state-of-the-art approaches.

\subsection{Decomposition principle}


We consider the problem of predicting, from a random variable $\X$, the random variable $\Y = f(\X)$, for a given function $f$.%
\footnote{Capital letters are reserved for random variables and small letters for their realisations.}
This function can usually be stochastic and is thought as any process that represent a physical or virtual relation between the two modalities $\X$ and $\Y$.
Once the function parameterised by a neural network $f_\theta$, where $\theta$ denotes the parameters of the network, we look for a meaningful description of the training procedures utilised in modern machine learning.
A natural consideration is to express the mutual information between $\X$ and $\Y$
\begin{align}
    I(\X;\Y)
    = \E_{p(\x,\y)}[\log\frac{p(\y|\x)}{p(\y)}],
    \label{eq:mi}
\end{align}
and take it as a maximum objective towards which we want to push our neural network.
To achieve this, we inject a parametric conditional density $p_\theta(\y|\x)$, characterised by the network outputs, and use the bound proposed in \cite{barber2004algorithm}
\begin{align}
\begin{split}
    I(\X;\Y)
    &= \E_{p(\x,\y)}[\log\frac{p_\theta(\y|\x)}{p(\y)}]
    + \E_{p(\x)}[\kld{p(\y|\x)}{p_\theta(\y|\x)\fphantom}] \\
    &\geq \E_{p(\x,\y)}[\log\frac{p_\theta(\y|\x)}{p(\y)}],
    \label{eq:ba_bound}
\end{split}
\end{align}
in order to get a parametric lower bound to maximise.
However, practical implementations of such a term typically require pairs of data $\{(\x_i,\y_i)\}_{i=1}^N$ in order to evaluate the only $\theta$-dependent part $\E_{p(\x,\y)}[\log \, p_\theta(\y|\x)] = -H(p(\y|\x);p_\theta(\y|\x))$, namely the negative conditional cross-entropy.
Although suitable for some simple datasets, the vast majority of modern problems involve unlabelled data, or could greatly benefit from its use.
To also exploit the unpaired data, similarly to the development described in \cite{quetant2023turbo}, we inject the parametric marginal density $p_\theta(\y)$ and re-express the bound as
\begin{align}
\begin{split}
    I(\X;\Y)
    &\geq \E_{p(\x,\y)}[\log\frac{p_\theta(\y|\x)}{p_\theta(\y)}]
    - \kld{p(\y)}{p_\theta(\y)\fphantom}.
    \label{eq:twin_bound}
\end{split}
\end{align}
This formulation has the benefit of involving an additional \textit{divergence} term between the real and parametric marginal densities, which can be used in certain situations to ensure better coverage of the data distribution.
In practice, this term is usually implemented as a discriminator classifier.
The other appearance of the parametric marginal density through the cross-entropy $-\E_{p(\y)}[\log \, p_\theta(\y)] = H(p(\y);p_\theta(\y))$ leads to so-called \textit{contrastive}-like training.
Indeed, it creates an opposition between the likelihood of samples from paired and unpaired data, often referred to as positive and negative pairs, respectively.
We can also neglect the cross-entropy $H(p(\y);p_\theta(\y))$ since it is non-negative, in principle for discrete random variable only, as presented in \cite{quetant2023turbo}.
This study argues that if it is appropriate to the problem at hand, it should be retained.


\subsection{Density parameterisations}


As a generic formula, any parametric conditional density $p_\theta(\y|\x)$ can be expressed as a carefully normalised exponential function such as
\begin{align}
    p_\theta(\y|\x) &= \frac{p(\y)}{Z_\theta(\x)} \, e^{s_\theta(\x,\y)}, && Z_\theta(\x) = \sum_{\y \in \mathcal{Y}} p(\y) \, e^{s_\theta(\x,\y)}, \label{eq:scalenorm_density} \\
    p_\theta(\y|\x) &= \frac{1}{Z_\theta(\x)} \, e^{s_\theta(\x,\y)}, && Z_\theta(\x) = \sum_{\y \in \mathcal{Y}} e^{s_\theta(\x,\y)}, \label{eq:globalnorm_density} \\
    p_\theta(\y|\x) &= e^{s_\theta(\x,\y)}, && 1 = \sum_{\y \in \mathcal{Y}} e^{s_\theta(\x,\y)}, \label{eq:selfnorm_density}
\end{align}
where $Z_\theta(\x)$ is the partition function and $s_\theta(\x,\y)$ is a score associated to the $(\x,\y)$ pair.
We call \cref{eq:scalenorm_density,eq:globalnorm_density,eq:selfnorm_density} the \textit{scaled-normalised}, \textit{globally-normalised} and \textit{self-normalised} density cases, respectively.
Note that the partition function of the scaled-normalised density can be expressed as $Z_\theta(\x) = \sum_{\y \in \mathcal{Y}} p(\y) \, e^{s_\theta(\x,\y)} = \E_{p(\y)}[e^{s_\theta(\x,\y)}]$.
With these expressions at hands, it is straightforward to define the corresponding marginal density as
\begin{equation}
    p_\theta(\y) = \E_{p(\x)}[p_\theta(\y|\x)].
    \label{eq:marginal_density}
\end{equation}

\subsection{Bound formulations}

The lower bound and density parameterisations presented so far are theoretical formulas valid in general.
In order to formulate proper loss functions, approximations are often necessary.
We present here the generic derivations of the objectives used in our experiments.
The translation into the actual setup and notations will be discussed later.

\paragraph{Categorical Cross-Entropy.}

The formalism leads to the popular categorical cross-entropy loss when a carefully chosen score function is used.
Let us define $s_\theta(\x,\y) = \y \cdot f_\theta(\x) = \sum_{c=1}^C y_c \, f_\theta(\x)_c$, where the vector-valued function $f_\theta$ is a neural network often called a classifier and where $y_c$ is the $c$-th component of the one-hot vector $\y \in \{0, 1\}^C$ indicating to which of the $C$ classes the data $\x$ belongs.
The normalisation of the globally-normalised density of \cref{eq:globalnorm_density} becomes the sum over the $C$ classes $Z_\theta(\x) = \sum_{\y \in \mathcal{Y}} e^{\, \y \cdot f_\theta(\x)} = \sum_{c=1}^C e^{\, f_\theta(\x)_c}$ and the conditional density finally becomes the \textit{softmax} function
\begin{align}
    p_\theta(\y|\x) = \frac{e^{\, \y \cdot f_\theta(\x)}}{\sum_{\y' \in \mathcal{Y}} e^{\, \y' \cdot f_\theta(\x)}}.
    \label{eq:softmax}
\end{align}
The role of the softmax function is to rescale the predictions of the classifier to values in the interval $(0, 1)$ and that sum to 1.
Softening the latter condition, we can use the self-normalised density parameterisation of \cref{eq:selfnorm_density} instead and replace the rescaling with the \textit{sigmoid} function
\begin{align}
    p_\theta(\y|\x) = \frac{e^{\, \y \cdot f_\theta(\x)}}{1 + e^{\, \y \cdot f_\theta(\x)}}.
    \label{eq:sigmoid}
\end{align}
We concisely denote both formulations as $p_\theta(\y|\x) = \sigma_{\y}(f_\theta(\x))$.

In most applications, categorical cross-entropy training problems only take into account the conditional term $\E_{p(\x,\y)}[\log \, p_\theta(\y|\x)] = -H(p(\y|\x);p_\theta(\y|\x))$.
In such a case, sigmoid rescaling leads to a collapse of all classifier outputs to 1 as a trivial maximum.
We argue in this study that a better choice is to use the tighter lower bound developed in \cref{eq:twin_bound}, leading to
\begin{align}
\begin{split}
    I(\X;\Y)
    &\geq \E_{p(\x,\y)}[\log\frac{\sigma_{\y}(f_\theta(\x))}{\E_{p(\x')}[\sigma_{\y}(f_\theta(\x'))]}]
    - \kld{p(\y)}{p_\theta(\y)\fphantom},
    \label{eq:cat_twin}
\end{split}
\end{align}
where the exact nature of the function $\sigma_{\y}$ is left open.
Both the denominator of the contrastive term and the divergence prevent the classifier from collapsing.

\paragraph{Binary Cross-Entropy.}

A potential drawback of this approach is that the log-sum-exp expansion (see \cref{app:logsumexp}) cannot be used to compute the denominator term $-\log\E_{p(\x')}[\sigma_{\y}(f_\theta(\x'))]$ efficiently, which can therefore be numerically unstable.
However, we observe that its role in the loss function is to force the classifier to output values closer to 0.
This contrasts with the numerator term $\log \sigma_{\y}(f_\theta(\x))$, which forces the output for the correct class towards 1.
A stable alternative that mimic this contrastive goal can be found in binary cross-entropy.
Indeed, the categorical cross-entropy formulation assumes that the predicted probability of the class $\y$ is fully given by the corresponding output of the classifier.
The binary cross-entropy formulation includes all the outputs and tries to push not only the correct one towards 1, but also the wrong ones towards 0.
In essence, this has a similar role as contrastive learning.
This is achieved by reformulating the conditional density as the product of the probability that each classifier output gives the correct answer
\begin{align}
    p_\theta(\y|\x)
    = \prod_{\substack{c=1\\y_c=1}}^C \left[\sigma(f_\theta(\x))_c\fphantom\right] \cdot \prod_{\substack{c=1\\y_c=0}}^C \left[1 - \sigma(f_\theta(\x))_c\fphantom\right],
    \label{eq:bin_density}
\end{align}
where $y_c$ is the $c$-th component of the one-hot vector $\y$ and where $\sigma$ (without subscript) is the vector-valued softmax or sigmoid function applied to the vector $f_\theta(\x)$.%
\footnote{The relation between $\sigma_{\y}$ and $\sigma$ can be understood as $\sigma_{\y}(f_\theta(\x)) = \y \cdot \sigma(f_\theta(\x))$ for a one-hot vector $\y$.}
Plugging this expression in \cref{eq:twin_bound} leads to the usual binary cross-entropy
\begin{align}
\begin{split}
    I(\X;\Y)
    &\geq \E_{p(\x,\y)}[\sum_{c=1}^C y_c \log \sigma(f_\theta(\x))_c]
    + \E_{p(\x,\y)}[\sum_{c=1}^C (1-y_c) \log(1-\sigma(f_\theta(\x))_c)] \\
    &- \kld{p(\y)}{p_\theta(\y)\fphantom},
    \label{eq:bin_twin}
\end{split}
\end{align}
where the neglected unstable denominator is compensated for by a conditional density designed to achieve the same objective.

\paragraph{InfoNCE.}

We develop here two slightly different ways to implement the InfoNCE loss.
By using our \twinturbo formalism of \cref{eq:twin_bound}, we can plug the self-normalised density of \cref{eq:selfnorm_density} to get
\begin{align}
\begin{split}
    I(\X;\Y)
    &\geq \E_{p(\x,\y)}[\log\frac{e^{s_\theta(\x,\y)}}{\E_{p(\x')}[e^{s_\theta(\x',\y)}]}]
    - \kld{p(\y)}{p_\theta(\y)\fphantom}.
    \label{eq:twinnce}
\end{split}
\end{align}
The choice of the exact form of the score $s_\theta(\x,\y)$ is free.
This formulation needs the divergence term in order to keep being a lower bound on the mutual information.
For adapted problems, this additional constraint on the global shape of the data manifold can be very helpful.
However, the score function then needs to be flexible enough to properly self-normalise.

If the dataset does not allow a proper calculation of the divergence, e.g. because there is no meaningful prior on the marginal density, another lower bound is required.
A simple way the recover the usual InfoNCE loss is to depart from the bound of \cref{eq:ba_bound} originally proposed in \cite{barber2004algorithm} and use the scaled-normalised density of \cref{eq:scalenorm_density}
\begin{align}
\begin{split}
    I(\X;\Y)
    &\geq \E_{p(\x,\y)}[\log\frac{\frac{p(\y)}{Z_\theta(\x)} \, e^{s_\theta(\x,\y)}}{p(\y)}]
    = \E_{p(\x,\y)}[\log\frac{e^{s_\theta(\x,\y)}}{\E_{p(\y')}[e^{s_\theta(\x,\y')}]}].
    \label{eq:infonce}
\end{split}
\end{align}
Note the subtle but crucial difference between the expectations in the denominator of \cref{eq:twinnce}, derived from the marginal density $p_\theta(\y)$, and the denominator of \cref{eq:infonce}, derived from the partition function $Z_\theta(\x)$.

\paragraph{Discriminator.}

A Kullback-Leibler divergence term is present in \cref{eq:cat_twin,eq:bin_twin,eq:twinnce}.
Unfortunately, this quantity is intractable in practice and usually replaced by the symmetric Jensen-Shannon divergence.
In order to compute it, the density ratio trick is often used through a discriminator network.
It follows from defining the ratio of densities as $r_\theta(\y) = \frac{p(\y)}{p_\theta(\y)} \equiv \frac{p_\alpha(\y|i=1)}{p_\alpha(\y|i=0)} = \frac{p_\alpha(i=1|\y)}{p_\alpha(i=0|\y)}$ where $p(i=0) = p(i=1)$ for an indicative variable $i$ and a set of parameters $\alpha$.%
\footnote{The $\theta$ dependency is effectively present through $\y = f_\theta(\x)$.}
By defining a discriminator network $D_\alpha(\y) := p_\alpha(i=1|\y)$, we can express the Jensen-Shannon divergence as
\begin{align}
\begin{split}
    \jsd{p(\y)}{p_\theta(\y)}
    &= \frac{1}{2} \E_{p(\y)}[\log(D_\alpha(\y))]
    + \frac{1}{2} \E_{p_\theta(\y)}[\log(1-D_\alpha(\y))]
    + \log 2,
    \label{eq:jsd}
\end{split}
\end{align}
which is the negative binary cross-entropy, up to the non-trainable $\log 2$ term and $1/2$ factor.
We can now alternatively train the discriminator $D_\alpha$ to maximise and the predictor $f_\theta$ to minimise \cref{eq:jsd}.
A perfectly trained discriminator would also allow to estimate the Kullback-Leibler divergence via
\begin{align}
\begin{split}
    \kld{p(\y)}{p_\theta(\y)}
    &= \E_{p(\y)}[\log(D_\alpha(\y))]
    - \E_{p(\y)}[\log(1-D_\alpha(\y))],
    \label{eq:kld}
\end{split}
\end{align}
but this expression does not depend on $\theta$ and cannot be used to train $f_\theta$.
Therefore, as is common practice, we use \cref{eq:jsd} as a proxy for the Kullback-Leibler divergence in \cref{eq:cat_twin,eq:bin_twin,eq:twinnce}.

\section{Experiments}
\label{sec:experiment}

\begin{wrapfigure}[24]{R}{0.59\textwidth}
    \includegraphics[scale=\diagramscale, trim={5 0 90 0}]{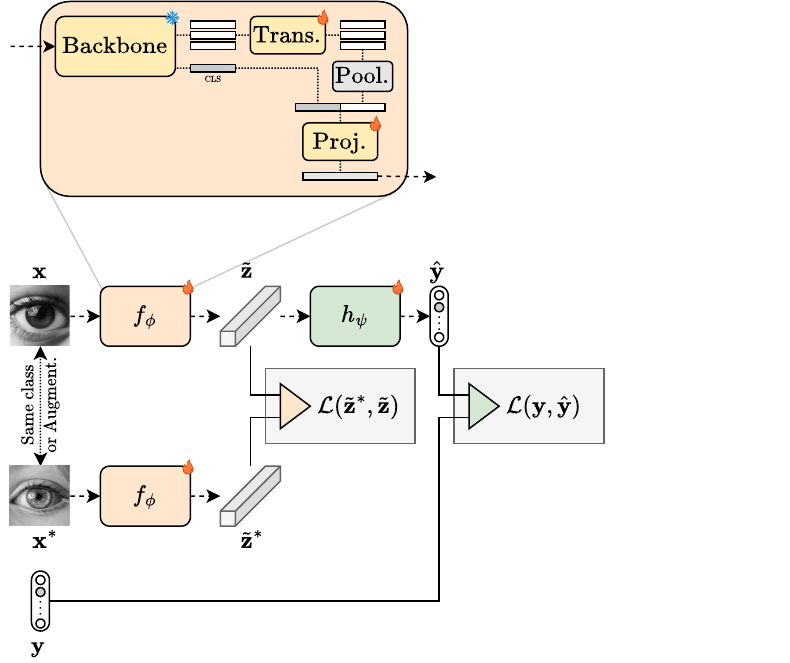}
    \caption{Model architecture and training setup.}
    \label{fig:architecture}
\end{wrapfigure}

We test our \twinturbo formalism on several semi-supervised classification experiments where the number of labelled data is kept below 0.2\% and 2\%.

\subsection{Datasets}

The datasets examined are MNIST~\cite{mnist}, CIFAR10~\cite{cifar} and SVHN~\cite{svhn}.
MNIST provides 60'000 greyscale handwritten digits from 0 to 9 and is considered to be an easy dataset to learn.
CIFAR10 is an analogous dataset that focuses on 50'000 natural images.
It contains 10 classes of low resolution photographs of animals and vehicles.
SVHN is a much harder dataset of 73'257 house numbers taken from the street.
Only the central digit of the image determines the class.
Therefore, it has a lot of contamination from surrounding numbers and background, making it a perfect benchmark for testing our framework.
All three datasets provide an additional test set of 10'000 samples.

The semi-supervised setup is constructed by creating a dual dataset that samples in parallel all images with labels removed and a small duplicated subset of 100 or 1000 labelled images.
This reflects a situation with very little labelled data.
The labelled samples are used to compute the supervised cross-entropy and latent regularisation losses, while the unlabelled samples are used by the discriminator and the self-supervised latent regularisation loss, as described below.

\subsection{Model architecture}

To fully exploit the dataset and the \twinturbo formalism, our fine-tuning model follows an encoder-predictor architecture as depicted in \cref{fig:architecture}.
The encoder $f_\phi$ consists of a frozen RADIOv2.5-B~\cite{ranzinger2024amradio,ranzinger2024phis,heinrich2024radioamplified} foundation model backbone, plus a trainable transformer and projector.%
\footnote{Only the RADIOv2.5-B is tested, as it shows state-of-the-art results in \cite{drozdova2024semisupervised} and a larger model would require too expensive computing capacity.
Furthermore, we focus on a theoretically grounded study and present a proof of principle.
It is straightforward to apply the same setup to any other foundation model.}
The backbone takes an image $\x$ as input, creates patches from it and outputs several patch tokens and a global representation, also called a CLS token.
Most classification experiments based on foundation model representations use only the CLS token, as it corresponds to the global features extracted from the input image.
However, there is no guarantee that it contains all the necessary information, as certain local details may still be useful to identify the content of an image.
Therefore, inspired by \cite{jose2024dinov2meetstext}, we also consider the patch tokens.
They are passed in a single layer transformer to allow cross-interactions before an average pooling operation is performed.%
\footnote{We do not pass the CLS token through the transformer layer because i) the backbone is assumed to have already extracted the required features from the patches and ii) the higher dimension of the CLS token relative to the patch tokens created by the RADIOv2.5-B model would require a transformer with many more trainable parameters.}
The resulting token is concatenated with the CLS token and a final projector produces the latent representation $\zt = f_\phi(\x)$.
The role of the second copy of the encoder as well as the $\x^*$ and $\zt^*$ variables in the latent space alignment will be explained later.
The predictor $h_\psi$ is a classifier head that predicts the class $\yh = h_\psi(\zt)$ of the input image from the latent representation $\zt$.
We emphasise again that the foundation model is frozen, even though for brevity we refer to the whole backbone-transformer-projector block as $f_\phi$.
This is also a hint to the seamless extension of the framework to a trainable backbone.
Our network architecture follows \cite{taran2020variationalibn,drozdova2024semisupervised}.
More details are available in \cref{app:architecture}.

\subsection{Losses}

The \twinturbo formalism is used in both the latent space $\Z$ and downstream task space $\Y$ of our setup.
We translate here the lower bounds on mutual information presented above into losses to be minimised in each of these spaces.

\paragraph{Supervised classification.}

In the downstream task space, the losses are derived from the mutual information $I(\X;\Y)$ between the random variables $\X$ and $\Y$.
We use three variants of the the cross-entropy derived in \cref{eq:cat_twin,eq:bin_twin}.
Considering a batch of $N$ independent samples $\{(\x_i,\y_i)\}_{i=1}^N$, the baseline categorical cross-entropy loss corresponds to the numerator of the first term of \cref{eq:cat_twin}
\begin{align}
    \L^\text{cat-cross}
    = -\frac{1}{N} \sum_{i=1}^N \log\sigma_{\y_i}(h_\psi(f_\phi((\x_i))),
\end{align}
the categorical \textit{twin-entropy} loss corresponds to the entire first term
\begin{align}
    \L^\text{cat-twin}
    = -\frac{1}{N} \sum_{i=1}^N \log\frac{\sigma_{\y_i}(h_\psi(f_\phi((\x_i)))}{\frac{1}{N} \sum_{j=1}^N \sigma_{\y_i}(h_\psi(f_\phi(\x_j')))},
\end{align}
and the binary cross-entropy loss corresponds to the first two terms of \cref{eq:bin_twin}
\begin{align}
\begin{split}
    \L^\text{bin-cross}
    &= -\frac{1}{N} \sum_{i=1}^N \sum_{c=1}^C \y_{i,c} \log \sigma(h_\psi(f_\phi(\x_i)))_c \\
    &- \frac{1}{N} \sum_{i=1}^N \sum_{c=1}^C (1-\y_{i,c}) \log(1-\sigma(h_\psi(f_\phi(\x_i)))_c),
\end{split}
\end{align}
where the sums over $i$ and $j$ are the empirical approximations to the true expectations over the batch of $N$ samples.
These losses use only the labelled data and force the network to predict the correct class $\y$ for the input image $\x$.

\paragraph{Unsupervised manifold covering.}

In order for the three expressions above to form proper bounds, we must not omit the Kullback-Leibler divergence term derived earlier.
As already stated, we use the density ratio trick in each case and compute
\begin{align}
\begin{split}
    \L^\text{critic}
    &= \frac{1}{N} \sum_{i=1}^N \log(D_\alpha(\y_i))
    + \frac{1}{N} \sum_{i=1}^N \log(1-D_\alpha(h_\psi(f_\phi(\x_i))))].
\end{split}
\end{align}
This loss ensures that the $p_\theta(\y)$ distribution of the model's output is consistent with the categorical distribution given by $p(\y)$.
The $f_\phi$ and $h_\psi$ networks are trained by minimising $\L^\text{critic}$, while the $D_\alpha$ discriminator is trained to maximise it.
To calculate this loss, the entire dataset is used, without labels.
Our discriminator architecture follows \cite{taran2020variationalibn,drozdova2024semisupervised}.
More details are available in \cref{app:architecture}.

\paragraph{Supervised and self-supervised latent alignment.}

In addition to the downstream task space, we can also improve the representation given by the backbone foundation model and processed by the transformer and projector.
To do this, we apply the InfoNCE bound of \cref{eq:infonce} on the mutual information $I(\X;\Z^*)$ between the $\X$ and $\Z^*$ spaces and obtain the loss
\begin{align}
    \L^\text{latent/augment.}
    = -\frac{1}{N} \sum_{i=1}^N \log\frac{e^{s_\phi(\x_i,\zt^*_i)}}{\frac{1}{N} \sum_{j=1}^N e^{s_\phi(\x_i,\zt^*_j)}},
\end{align}
where the score function $s_\phi(\x,\zt^*) = \frac{\zt^* \cdot f_\phi(\x)}{\lVert \zt^* \rVert \lVert f_\phi(\x) \rVert}$ is chosen to be the cosine similarity.
A crucial aspect here is that the usual \textit{positive} and \textit{negative} pairs can be created in two different ways.
On the one hand, the common self-supervised creation of pairs is done via augmentations of the input images, passed through the network to create a new latent representation $\zt^* = f_\phi(\x^*)$.
The entire unlabelled training dataset can be used in this case and the loss is denoted as $\L^\text{augment.}$.
On the other hand, the supervised pairs are not created by artificially augmenting the input image, but rather by sampling another image from the same or a different class, respectively.
This is possible because of the labelled nature of the data used in this space through the $\L^\text{latent}$ loss.
In both cases, the variable $\zt^* = f_\phi(\x^*)$ is assumed to be the ground truth to which the encoder output $\zt = f_\phi(\x)$ must be aligned.
Note that the same network is used to generated both latent variables, without any update of its parameters.
More details are available in \cref{app:architecture}.

\paragraph{Combined loss.}

The combined total loss used to train the entire model is
\begin{align}
    \L = \L^\text{cat-cross/cat-twin/bin-cross} + \lambda_C \, \L^\text{critic} + \lambda_L \, \L^\text{latent} + \lambda_A \, \L^\text{augment.},
    \label{eq:loss_total}
\end{align}
where $\lambda_C$, $\lambda_L$ and $\lambda_A$ are weight hyperparameters that trade-off the different loss components.
In the following these weights and their corresponding loss terms will be referred to as \textit{critic}, \textit{latent} and \textit{augmentation} weights or losses, respectively.
In the legend of the figures, they are simply abbreviated as \textit{critic}, \textit{latent} and \textit{augment}.
It is important to note that we always train the entire model on the total loss.
No individual scheduler of the different weights is used, which means that every term of \cref{eq:loss_total} is activated all the time, except when its respective weight is equal to 0.

\subsection{Results}

\begin{figure}[tb]
\centering
\begin{subfigure}[t]{0.49\textwidth}
    \centering
    \includesvg[scale=0.42]{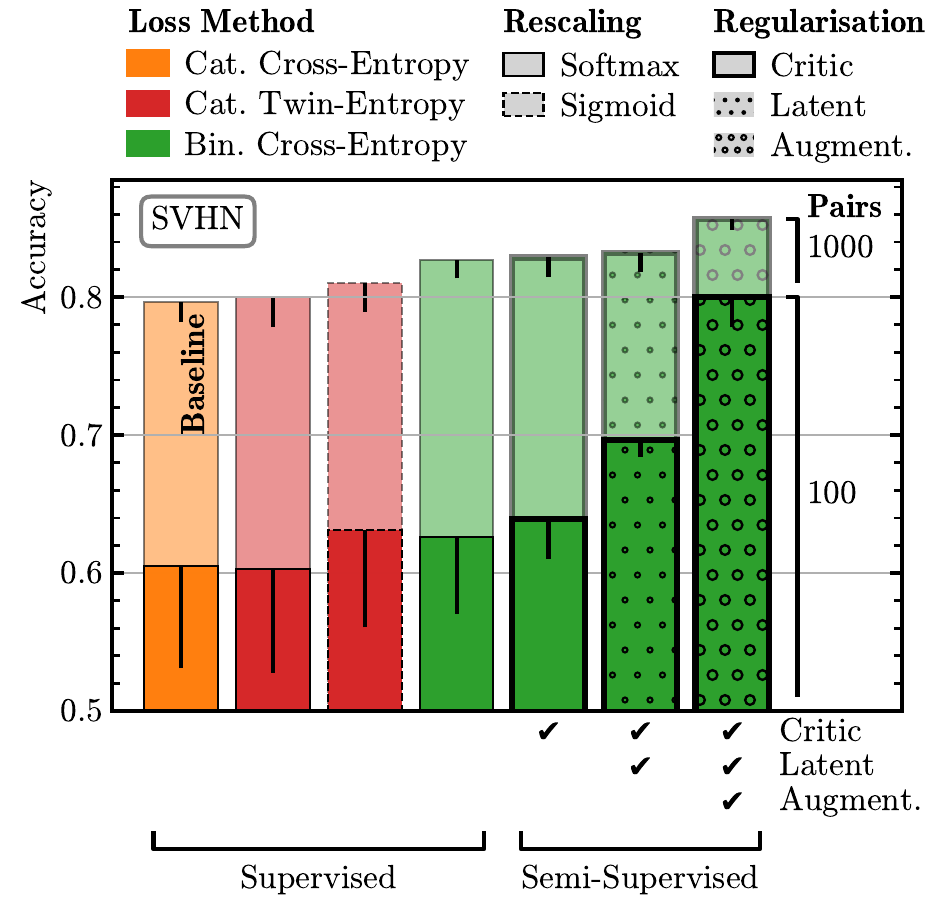}
\end{subfigure}%
\hfill
\begin{subfigure}[t]{0.49\textwidth}
    \centering
    \includesvg[scale=0.42]{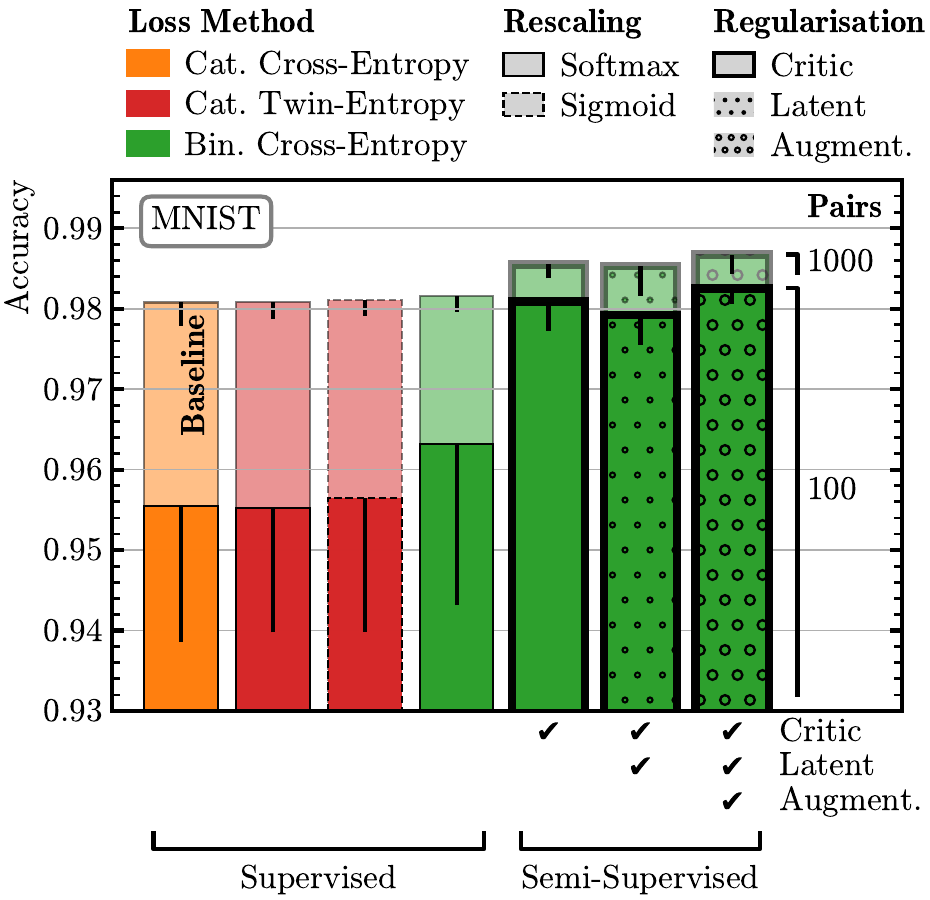}
\end{subfigure}%
\caption{
Ablation study of the classification accuracy for the SVHN (left) and MNIST (right) datasets.
The baseline training uses categorical cross-entropy loss $\L^\text{cat-cross}$ only with the softmax rescaling.
For SVHN, the number of unlabelled samples is 73'257 and the weights are set to $\lambda_C = 0.001$, $\lambda_L = 0.1$ and $\lambda_A = 0.1$.
For MNIST, the number of unlabelled samples is 60'000 and the weights are set to $\lambda_C = 1.0$, $\lambda_L = 0.1$ and $\lambda_A = 0.1$.
}
\label{fig:ablation}
\end{figure}

We present an ablation study of our framework in \cref{fig:ablation} for the SVHN and MNIST datasets.
The results for the CIFAR10 dataset are available in \cref{app:plots}.
The figures show the evolution of the classification accuracy, estimated on the test set, as each term of \cref{eq:loss_total} is activated one by one.
Both cases of 100 and 1000 labelled images are shown, denoted by the number of pairs in the dataset.
We also test the difference between the softmax and sigmoid rescaling functions of \cref{eq:softmax,eq:sigmoid}, respectively.
The \textit{baseline} represents the standard method typically used to fine-tune foundation models in a supervised manner.
The other bars show how performance evolves when our semi-supervised framework is fully leveraged.
In each case, the $\lambda_C$, $\lambda_L$ and $\lambda_A$ weights are fixed to their overall best performing values.
Moreover, each setup is trained three times with three different seeds.%
\footnote{Seeds are 42, 1337 and 3435.}
The best seed is selected and the range of accuracy achieved is shown as a vertical line on each bar, highlighting the variability of the setup.
Exhaustive results of the whole sweep over the weights are shown in \cref{app:plots}.

The results obtained for SVHN and shown in \cref{fig:ablation} (left) confirm that, as expected, the performance of the model is greatly improved when more pairs are available for the supervised losses.
This is reflected in a 20\% increase in accuracy in the baseline case.
Intriguingly, switching to the categorical twin-entropy loss does not change the results much if we stick to the softmax rescaling.
However, also switching to the sigmoid rescaling adds 1\% to 3\% accuracy, supporting our claim that the tighter bound of \cref{eq:twin_bound} has a positive impact on the convergence of the model.
We believe that relaxing the constraints on the sum of the predictor outputs allows the network to be more expressive, so that the contrastive-like denominator of $\L^\text{cat-twin}$ is able to guide the training towards a better local minimum of the loss.
We observe another 2\% boost in performance in the case of 1000 pairs when we switch to the more stable binary cross-entropy alternative to our contrastive-like loss.%
\footnote{At this point, sigmoid rescaling shows a decrease in performance, requiring deeper investigation.}
These results confirms both that supervised contrastive training leads to a gain in accuracy and that $\L^\text{bin-cross}$ is indeed more stable than $\L^\text{cat-twin}$ in its current implementation.
The results discussed so far only employed supervised loss in the downstream task space.
Adding the unsupervised critic loss does slightly improve the accuracy on the SVHN dataset, but more interestingly decreases the variability over seeds.
In other words, it stabilises the training.
This behaviour, also studied in \cite{taran2020variationalibn}, shows that an unsupervised regularisation of the output space, exploiting the large amount of unlabelled data, is necessary in the situation of low-labelled regimes.
A further 6\% improvement, for the case of 100 pairs, is due to the activation of the supervised latent alignment loss.
Having a more consistent latent space with respect to the different classes presumably favours a better prediction.
Finally, including unlabelled data in the latent space alignment loss by creating pairs via augmentations results in a significant improvement in performance.
The increase in accuracy is approximately 3\% and 10\% for the cases of 1000 and 100 pairs, respectively, which greatly reduces the gap between them.
This last point is particularly worth emphasising: by fully exploiting the unlabelled data and all the consistent loss terms derived in our framework, the model performs nearly as well with 100 and 1000 pairs.
Another important observation is that variability across seeds is reduced as soon as semi-supervised training is activated, with this effect becoming even more pronounced when the latent space is regularised.

Qualitatively, similar results are observed for the MNIST dataset in \cref{fig:ablation} (right).
However, the overall accuracy is much higher in this case due to the much simpler nature of the dataset and probably because these images have already been seen by the backbone foundation model during its pre-training phase.
Some notable differences are worth mentioning.
We observe that activating the critic loss is highly effective, almost bridging the gap between the cases of 100 and 1000 pairs.
We believe that, due to the aforementioned simplicity of the dataset and the high accuracy achieved, the impact of supervised classification loss may be reduced.
Therefore, we expect the additional unsupervised constraints to have a more visible effect.
It then seems that the performance reaches a plateau, as the accuracy is only marginally impacted by both supervised and self-supervised latent alignment losses.

Overall, we find that the performance of the classifier is highly dependent on the dataset, and most likely on what has already been seen during the pre-training of the foundation model.
We can assume that samples from MNIST and CIFAR10, two very common datasets in the machine learning community, were used to train the foundation model.
Therefore, our hypothesis is that the goal of the fine-tuning strategy is to retrieve these samples from the model's memory so that it starts working properly for classification.
This study shows that mutual information maximisation, and in particular the lower bound derived in \cref{eq:twin_bound}, works better towards this goal than the naive supervised categorical cross-entropy alone, especially for low-labelled data regimes.

\section{Conclusion}

We demonstrate that our mutual information maximisation framework is able to enhance the fine-tuning of foundation model-based classifiers thanks to three crucial aspects.
Firstly, although it is possible to form a lower bound that includes only the conditional density as in \cref{eq:ba_bound}, it is preferable to decompose it further and let the Kullback-Leibler divergence appear as in \cref{eq:twin_bound} in situations where a prior on the marginal density is known.
Note that in such cases this divergence term should not be omitted in order for the expression to be a proper lower bound.
%
Secondly, the appearance of the Kullback-Leibler divergence is closely related to the emergence of the contrastive-like denominator in \cref{eq:twin_bound}.
Preventing the predictor outputs from collapsing, this term allows the use of the more flexible sigmoid rescaling.
It also leads to an improvement in performance by leveraging the relationships between each pair of samples in a batch of data.
Thirdly, we observe that replacing our categorical twin-entropy loss with a binary cross-entropy stable alternative, which plays an analogous role, gives a new boost to the fine-tuning of the model.
%
Finally, the supervised and self-supervised latent space alignment losses, also expressed as lower bounds on the mutual information, bring a further increase in accuracy.
Indeed, the more the latent representations of images from the same class coincide, the better the predictor can rely on them to classify the images.
The results on MNIST, CIFAR10 and SVHN undeniably support the valorisation of the available unlabelled data.

All of these improvements are even more pronounced when we have only a few labelled samples at our disposal.
This can be particularly relevant in various situations, such as in scientific domains, where many observations are made, but only a hand-picked selection is accurately labelled due to time constraints, noisy data, unknown variables, etc.
Our framework would fit well in such situations.
In addition, while the focus of this study is on vision models, the proposed methods hold promise for extension to multimodal models, such as vision-language models, paving the way for broader applications in semi-supervised learning.
As a final thought, our results suggest that foundation models should be trained on as many datasets as possible in order to enable efficient fine-tuning.
Moreover, only a proper fine-tuning method would benefit from this and fully exploit the memory retrieval capability of such a foundation model.

\bibliography{bib/biblio}

\begin{thebibliography}{31}
\providecommand{\natexlab}[1]{#1}
\providecommand{\url}[1]{\texttt{#1}}
\expandafter\ifx\csname urlstyle\endcsname\relax
  \providecommand{\doi}[1]{doi: #1}\else
  \providecommand{\doi}{doi: \begingroup \urlstyle{rm}\Url}\fi

\bibitem[Voloshynovskiy et~al.(2020)Voloshynovskiy, Taran, Kondah, Holotyak, and Rezende]{taran2020variationalibn}
Slava Voloshynovskiy, Olga Taran, Mouad Kondah, Taras Holotyak, and Danilo Rezende.
\newblock {V}ariational {I}nformation {B}ottleneck for {S}emi-{S}upervised {C}lassification.
\newblock \emph{Entropy}, 22\penalty0 (9):\penalty0 943, 2020.
\newblock \doi{10.3390/e22090943}.

\bibitem[Quétant et~al.(2023)Quétant, Belousov, Kinakh, and Voloshynovskiy]{quetant2023turbo}
Guillaume Quétant, Yury Belousov, Vitaliy Kinakh, and Slava Voloshynovskiy.
\newblock {TURBO}: {T}he {S}wiss {K}nife of {A}uto-{E}ncoders.
\newblock \emph{Entropy}, 25\penalty0 (10):\penalty0 1471, 2023.
\newblock \doi{10.3390/e25101471}.

\bibitem[Drozdova et~al.(2024)Drozdova, Kinakh, Belousov, Lastufka, and Voloshynovskiy]{drozdova2024semisupervised}
Mariia Drozdova, Vitaliy Kinakh, Yury Belousov, Erica Lastufka, and Slava Voloshynovskiy.
\newblock {S}emi-{S}upervised {F}ine-{T}uning of {V}ision {F}oundation {M}odels with {C}ontent-{S}tyle {D}ecomposition, 2024.
\newblock URL \url{https://arxiv.org/abs/2410.02069}.

\bibitem[van~den Oord et~al.(2018)van~den Oord, Li, and Vinyals]{oord2018infonce}
Aaron van~den Oord, Yazhe Li, and Oriol Vinyals.
\newblock {Representation Learning with Contrastive Predictive Coding}, 2018.
\newblock URL \url{https://arxiv.org/abs/1807.03748}.

\bibitem[Chen et~al.(2020)Chen, Kornblith, Norouzi, and Hinton]{chen2020simclr}
Ting Chen, Simon Kornblith, Mohammad Norouzi, and Geoffrey Hinton.
\newblock A simple framework for contrastive learning of visual representations, 2020.
\newblock URL \url{https://arxiv.org/abs/2002.05709}.

\bibitem[Caron et~al.(2021)Caron, Touvron, Misra, J\'egou, Mairal, Bojanowski, and Joulin]{caron2021dino}
Mathilde Caron, Hugo Touvron, Ishan Misra, Herv\'e J\'egou, Julien Mairal, Piotr Bojanowski, and Armand Joulin.
\newblock {Emerging Properties in Self-Supervised Vision Transformers}.
\newblock In \emph{Proceedings of the IEEE/CVF International Conference on Computer Vision (ICCV)}, pages 9650--9660, 10 2021.

\bibitem[Radford et~al.(2021)Radford, Kim, Hallacy, Ramesh, Goh, Agarwal, Sastry, Askell, Mishkin, Clark, Krueger, and Sutskever]{radford2021clip}
Alec Radford, Jong~Wook Kim, Chris Hallacy, Aditya Ramesh, Gabriel Goh, Sandhini Agarwal, Girish Sastry, Amanda Askell, Pamela Mishkin, Jack Clark, Gretchen Krueger, and Ilya Sutskever.
\newblock {Learning Transferable Visual Models From Natural Language Supervision}, 2021.
\newblock URL \url{https://arxiv.org/abs/2103.00020}.

\bibitem[Lee(2013)]{lee2013pseudo}
Dong-Hyun Lee.
\newblock {P}seudo-{L}abel: {T}he {S}imple and {E}fficient {S}emi-{S}upervised {L}earning {M}ethod for {D}eep {N}eural {N}etworks.
\newblock \emph{Workshop on challenges in representation learning, ICML}, 2013.

\bibitem[Sajjadi et~al.(2016)Sajjadi, Javanmardi, and Tasdizen]{sajjadi2016regularization}
Mehdi Sajjadi, Mehran Javanmardi, and Tolga Tasdizen.
\newblock {R}egularization {W}ith {S}tochastic {T}ransformations and {P}erturbations for {D}eep {S}emi-{S}upervised {L}earning.
\newblock \emph{Conference on Neural Information Processing Systems}, 2016.

\bibitem[Laine and Aila(2017)]{laine2017temporal}
Samuli Laine and Timo Aila.
\newblock {T}emporal {E}nsembling for {S}emi-{S}upervised {L}earning.
\newblock \emph{International Conference on Learning Representations}, 2017.

\bibitem[Grandvalet and Bengio(2005)]{grandvalet2005semi}
Yves Grandvalet and Yoshua Bengio.
\newblock {S}emi-supervised {L}earning by {E}ntropy {M}inimization.
\newblock \emph{Conference on Neural Information Processing Systems}, 2005.

\bibitem[Sohn et~al.(2020)Sohn, Berthelot, Carlini, Zhang, Cubuk, Kurakin, Zhang, and Raffel]{sohn2020fixmatch}
Kihyuk Sohn, David Berthelot, Nicholas Carlini, Zizhao Zhang, Ekin Cubuk, Alex Kurakin, Han Zhang, and Colin Raffel.
\newblock {F}ix{M}atch: {S}implifying {S}emi-{S}upervised {L}earning with {C}onsistency and {C}onfidence.
\newblock \emph{Conference on Neural Information Processing Systems}, 2020.

\bibitem[Zhang et~al.(2021)Zhang, Wang, Hou, Wu, Wu, Zhu, Li, and Qi]{zhang2021flexmatch}
Bowen Zhang, Yidong Wang, Wenxin Hou, Qizhao Wu, Hao Wu, Zihan Zhu, Xuelong Li, and Guo-Jun Qi.
\newblock {F}lex{M}atch: {B}oosting {S}emi-{S}upervised {L}earning with {C}urriculum {P}seudo {L}abeling.
\newblock \emph{Conference on Neural Information Processing Systems}, 2021.

\bibitem[Hu et~al.(2021)Hu, Shen, Wallis, Allen-Zhu, Li, Wang, Wang, and Chen]{hu2021lora}
Edward~J. Hu, Yelong Shen, Phillip Wallis, Zeyuan Allen-Zhu, Yuanzhi Li, Shean Wang, Lu~Wang, and Weizhu Chen.
\newblock {LoRA}: {L}ow-{R}ank {A}daptation of {L}arge {L}anguage {M}odels, 2021.
\newblock URL \url{https://arxiv.org/abs/2106.09685}.

\bibitem[Kopiczko et~al.(2023)Kopiczko, Blankevoort, and Asano]{kopiczko2023vera}
Dawid~J. Kopiczko, Tijmen Blankevoort, and Yuki~M. Asano.
\newblock {VeRA}: {V}ector-based {R}andom {M}atrix {A}daptation, 2023.
\newblock URL \url{https://arxiv.org/abs/2310.11454}.

\bibitem[Mao et~al.(2024)Mao, Huang, Guan, Bao, Mo, and Xu]{mao2024dora}
Yulong Mao, Kaiyu Huang, Changhao Guan, Ganglin Bao, Fengran Mo, and Jinan Xu.
\newblock {DoRA}: {E}nhancing {P}arameter-{E}fficient {F}ine-{T}uning with {D}ynamic {R}ank {D}istribution, 2024.
\newblock URL \url{https://arxiv.org/abs/2405.17357}.

\bibitem[Mahabadi et~al.(2021)Mahabadi, Ruder, Dehghani, and Gurevych]{mahabadi2021compacter}
Rabeeh~Karimi Mahabadi, Sebastian Ruder, Mostafa Dehghani, and Iryna Gurevych.
\newblock {C}ompacter: {E}fficient {L}ow-{R}ank {H}ypercomplex {A}dapter {L}ayers.
\newblock \emph{Conference on Neural Information Processing Systems}, 2021.

\bibitem[Lester et~al.(2021)Lester, Al-Rfou, and Constant]{lester2021power}
Brian Lester, Rami Al-Rfou, and Noah Constant.
\newblock {The Power of Scale for Parameter-Efficient Prompt Tuning}.
\newblock \emph{Conference on Empirical Methods in Natural Language Processing}, 2021.

\bibitem[Tishby et~al.(2000)Tishby, Pereira, and Bialek]{tishby2000information}
Naftali Tishby, Fernando~C. Pereira, and William Bialek.
\newblock The information bottleneck method, 2000.
\newblock URL \url{https://arxiv.org/abs/physics/0004057}.

\bibitem[Barber and Agakov(2004)]{barber2004algorithm}
David Barber and Felix Agakov.
\newblock {T}he {IM} {A}lgorithm: {A} variational approach to {I}nformation {M}aximization.
\newblock \emph{Advances in neural information processing systems}, 16\penalty0 (320):\penalty0 201, 2004.

\bibitem[Alemi et~al.(2016)Alemi, Fischer, Dillon, and Murphy]{alemi2016deepvariational}
Alexander~A. Alemi, Ian Fischer, Joshua~V. Dillon, and Kevin Murphy.
\newblock {Deep Variational Information Bottleneck}, 2016.
\newblock URL \url{https://arxiv.org/abs/1612.00410}.

\bibitem[Poole et~al.(2019)Poole, Ozair, Van Den~Oord, Alemi, and Tucker]{poole2019variational}
Ben Poole, Sherjil Ozair, Aaron Van Den~Oord, Alex Alemi, and George Tucker.
\newblock On {V}ariational {B}ounds of {M}utual {I}nformation.
\newblock In \emph{International Conference on Machine Learning}, pages 5171--5180. PMLR, 2019.

\bibitem[Tschannen et~al.(2019)Tschannen, Djolonga, Rubenstein, Gelly, and Lucic]{tschannen2019mutualinformation}
Michael Tschannen, Josip Djolonga, Paul~K. Rubenstein, Sylvain Gelly, and Mario Lucic.
\newblock {O}n {M}utual {I}nformation {M}aximization for {R}epresentation {L}earning, 2019.
\newblock URL \url{https://arxiv.org/abs/1907.13625}.

\bibitem[Hjelm et~al.(2018)Hjelm, Fedorov, Lavoie-Marchildon, Grewal, Bachman, Trischler, and Bengio]{hjelm2018deepinfomax}
R~Devon Hjelm, Alex Fedorov, Samuel Lavoie-Marchildon, Karan Grewal, Phil Bachman, Adam Trischler, and Yoshua Bengio.
\newblock {L}earning deep representations by mutual information estimation and maximization, 2018.
\newblock URL \url{https://arxiv.org/abs/1808.06670}.

\bibitem[Lecun et~al.(1998)Lecun, Bottou, Bengio, and Haffner]{mnist}
Y.~Lecun, L.~Bottou, Y.~Bengio, and P.~Haffner.
\newblock {G}radient-based learning applied to document recognition.
\newblock \emph{Proceedings of the IEEE}, 86\penalty0 (11):\penalty0 2278--2324, 1998.
\newblock \doi{10.1109/5.726791}.

\bibitem[Krizhevsky(2009)]{cifar}
Alex Krizhevsky.
\newblock {L}earning {M}ultiple {L}ayers of {F}eatures from {T}iny {I}mages.
\newblock Technical report, 2009.
\newblock URL \url{https://www.cs.toronto.edu/~kriz/cifar.html}.

\bibitem[Netzer et~al.(2011)Netzer, Wang, Coates, Bissacco, Wu, Ng, et~al.]{svhn}
Yuval Netzer, Tao Wang, Adam Coates, Alessandro Bissacco, Baolin Wu, Andrew~Y Ng, et~al.
\newblock {R}eading {D}igits in {N}atural {I}mages with {U}nsupervised {F}eature {L}earning.
\newblock In \emph{Workshop on deep learning and unsupervised feature learning, NIPS}, volume 2011, page~4. Granada, 2011.

\bibitem[Ranzinger et~al.(2024{\natexlab{a}})Ranzinger, Heinrich, Kautz, and Molchanov]{ranzinger2024amradio}
Mike Ranzinger, Greg Heinrich, Jan Kautz, and Pavlo Molchanov.
\newblock {AM-RADIO}: {A}gglomerative {V}ision {F}oundation {M}odel {R}educe {A}ll {D}omains {I}nto {O}ne.
\newblock In \emph{Proceedings of the IEEE/CVF Conference on Computer Vision and Pattern Recognition (CVPR)}, pages 12490--12500, 6 2024{\natexlab{a}}.

\bibitem[Ranzinger et~al.(2024{\natexlab{b}})Ranzinger, Barker, Heinrich, Molchanov, Catanzaro, and Tao]{ranzinger2024phis}
Mike Ranzinger, Jon Barker, Greg Heinrich, Pavlo Molchanov, Bryan Catanzaro, and Andrew Tao.
\newblock {PHI-S}: {D}istribution {B}alancing for {L}abel-{F}ree {M}ulti-{T}eacher {D}istillation, 2024{\natexlab{b}}.
\newblock URL \url{https://arxiv.org/abs/2410.01680}.

\bibitem[Heinrich et~al.(2024)Heinrich, Ranzinger, Hongxu, Yin, Lu, Kautz, Tao, Catanzaro, and Molchanov]{heinrich2024radioamplified}
Greg Heinrich, Mike Ranzinger, Hongxu, Yin, Yao Lu, Jan Kautz, Andrew Tao, Bryan Catanzaro, and Pavlo Molchanov.
\newblock {RADIO} {A}mplified: {I}mproved {B}aselines for {A}gglomerative {V}ision {F}oundation {M}odels, 2024.
\newblock URL \url{https://arxiv.org/abs/2412.07679}.

\bibitem[Jose et~al.(2024)Jose, Moutakanni, Kang, Baldassarre, Darcet, Xu, Li, Szafraniec, Ramamonjisoa, Oquab, Siméoni, Vo, Labatut, and Bojanowski]{jose2024dinov2meetstext}
Cijo Jose, Théo Moutakanni, Dahyun Kang, Federico Baldassarre, Timothée Darcet, Hu~Xu, Daniel Li, Marc Szafraniec, Michaël Ramamonjisoa, Maxime Oquab, Oriane Siméoni, Huy~V. Vo, Patrick Labatut, and Piotr Bojanowski.
\newblock {DINOv2} {M}eets {T}ext: {A} {U}nified {F}ramework for {I}mage- and {P}ixel-{L}evel {V}ision-{L}anguage {A}lignment, 2024.
\newblock URL \url{https://arxiv.org/abs/2412.16334}.

\end{thebibliography}
\bibliographystyle{unsrtnat}

\newpage
\appendix
\section{Log-sum-exp expansion}
\label{app:logsumexp}

The numerical instability of the softmax and sigmoid functions is a well-known problem.
Indeed, they can quickly underflow to zero or overflow to infinity, destroying the ability to train a network.
Fortunately, in many cases only the logarithm of these functions is present in the loss, and simple tricks can be used to stabilise the training.

To simplify the notation, let us define the \textit{softmax} function as
\begin{align}
    \sigma(\x)_i = \frac{e^{\x_i}}{\sum_i e^{\x_i}},
\end{align}
for a vector $\x$.
The numerator $e^{\x_i}$ can easily overflow or underflow due to the rapid evolution of the exponential function and the limited precision of a computer.
To solve the overflow issue, we can shift the exponent by computing equivalently
\begin{align}
    \sigma(\x)_i = \frac{e^{\x_i - \max(\x)}}{\sum_i e^{\x_i - \max(\x)}},
\end{align}
so that the exponential is always between 0 and 1.
This also ensures that the denominator is never 0, since at least one term of the sum is $e^0 = 1$.
To further solve the underflow issue when computing the logarithm of the softmax, we simply need to rewrite the formula as
\begin{align}
    \log\sigma(\x)_i = \x_i - \max(\x) - \log\sum_i e^{\x_i - \max(\x)},
\end{align}

The stabilisation of the \textit{sigmoid} function follows another simple trick.
Let us define it again as
\begin{align}
    \sigma(\x)_i = \frac{e^{\x_i}}{1 + e^{\x_i}}.
\end{align}
Taking the logarithm and rewriting the expression as
\begin{align}
\begin{split}
    \log\sigma(\x)_i
    &= \log\frac{e^{\x_i}}{1 + e^{\x_i}} \\
    &= \log\frac{1}{1 + e^{-\x_i}} \\
    &= -\log(1 + e^{-\x_i}),
\end{split}
\end{align}
solves the overflow issue.
To solve the underflow issue, we switch to the linear function when $\x_i \ll 0$, namely when $-\log(1 + e^{-\x_i}) \approx-\log(e^{-\x_i}) = \x_i$.

\section{Architecture details}
\label{app:architecture}

For the common networks, our model follows the architecture described in \cite{taran2020variationalibn,drozdova2024semisupervised}.
We highlight the relevant features of the RADIOv2.5-B model in \cref{tab:archi_radio} and summarise all the details of the transformer, projector, predictor and discriminator networks in \cref{tab:archi_trans,tab:archi_proj,tab:archi_pred,tab:archi_disc}, respectively.
When applied, augmentations involve resizing, cropping, horizontal flip (if not altering the semantic content of the image), colour jitter, grey scale, Gaussian blur and solarisation.%
\footnote{Parameters follow \url{https://github.com/facebookresearch/dinov2}.}
The models are trained with the AdamW optimiser for 5 epochs (around 500-700 steps) with a batch size of 512, a learning rate of $5 \cdot 10^{-5}$, and a 150-step linear warm-up scheduler with an initial factor of 0.001.
All models are trained on internal clusters of GPUs including NVIDIA A100, L40, GeForce RTX 3090 and RTX A5000/A5500 for approximately 1 to 2 hours each.

\subsection{Backbone}

The RADIOv2.5-B foundation model takes an image as input.
It outputs tokens corresponding to multiple patches of the image and a CLS token containing the global information about the image.
A few of its relevant features are highlighted in \cref{tab:archi_radio}.

\begin{table}[H]
    \centering
    \caption{Architecture highlights of the RADIOv2.5-B model.}
    \begin{tabular}{ccc}
        \toprule
        \textbf{Input} & \textbf{Output} & \textbf{Layer} \\
        \midrule
        28 $\times$ 28 $|$ 32 $\times$ 32 & 224 $\times$ 244 & Bilinear interpolation \\
         &  & \vdots \\
        / & 2304 & CLS token \\
        / & 768 & Patch tokens \\
        \bottomrule
    \end{tabular}
    \label{tab:archi_radio}
\end{table}

\subsection{Transformer}

The transformer takes the backbone patch tokens as input.
It applies a single layer with both a single-headed self-attention block and a SwiGLU network to allow further cross-interaction of the tokens.
Its architecture is detailed in \cref{tab:archi_trans}.

\begin{table}[H]
    \centering
    \caption{Architecture details of the transformer network.}
    \begin{tabular}{ccc}
        \toprule
        \textbf{Input} & \textbf{Output} & \textbf{Layer} \\
        \midrule
        768 & 768 & Single-Headed Self-Attention \\
         &  & \textbf{SwiGLU network} \\
        \midrule
        768 & 768 + 768 & Linear (chunks: 2) \\
        768 & 768 & SiLU (on chunk 1) \\
        768 + 768 & 768 & Mult(chunk 1, chunk 2) \\
        768 & 768 & Linear \\
        \bottomrule
    \end{tabular}
    \label{tab:archi_trans}
\end{table}

\subsection{Projector}

The projector takes the concatenation of the backbone CLS token and the average pooling of the transformer patch tokens as input.
It applies a dense network with dropout to project these features to a latent space.
Its architecture is detailed in \cref{tab:archi_proj}.

\begin{table}[H]
    \centering
    \caption{Architecture details of the projector network.}
    \begin{tabular}{ccc}
        \toprule
        \textbf{Input} & \textbf{Output} & \textbf{Layer} \\
        \midrule
        2304 + 768 & 8000 & Linear \\
        8000 & 8000 & LeakyReLU (slope: 0.01) \\
        8000 & 8000 & Dropout (probability: 0.3) \\
        \bottomrule
    \end{tabular}
    \label{tab:archi_proj}
\end{table}

\subsection{Predictor}

The predictor takes the projector latent representation as input.
It applies a dense network to classify the initial image by means of a one-hot output vector.
Its architecture is detailed in \cref{tab:archi_pred}.

\begin{table}[H]
    \centering
    \caption{Architecture details of the predictor network.}
    \begin{tabular}{ccc}
        \toprule
        \textbf{Input} & \textbf{Output} & \textbf{Layer} \\
        \midrule
        8000 & 1024 & Linear \\
        1024 & 1024 & LeakyReLU (slope: 0.01) \\
        1024 & 10 & Linear \\
        \bottomrule
    \end{tabular}
    \label{tab:archi_pred}
\end{table}

\subsection{Discriminator}

The discriminator takes the predictor one-hot vector as input.
It applies a dense network to predict whether this vector is consistent with a categorical distribution or not.
Its architecture is detailed in \cref{tab:archi_disc}.

\begin{table}[H]
    \centering
    \caption{Architecture details of the discriminator network.}
    \begin{tabular}{ccc}
        \toprule
        \textbf{Input} & \textbf{Output} & \textbf{Layer} \\
        \midrule
        10 & 500 & Linear \\
        500 & 500 & ReLU \\
        500 & 500 & Linear \\
        500 & 500 & ReLU \\
        500 & 1 & Linear \\
        \bottomrule
    \end{tabular}
    \label{tab:archi_disc}
\end{table}

\section{Additional plots}
\label{app:plots}

\begin{figure}[H]
    \centering
    \includegraphics[scale=\plotscale]{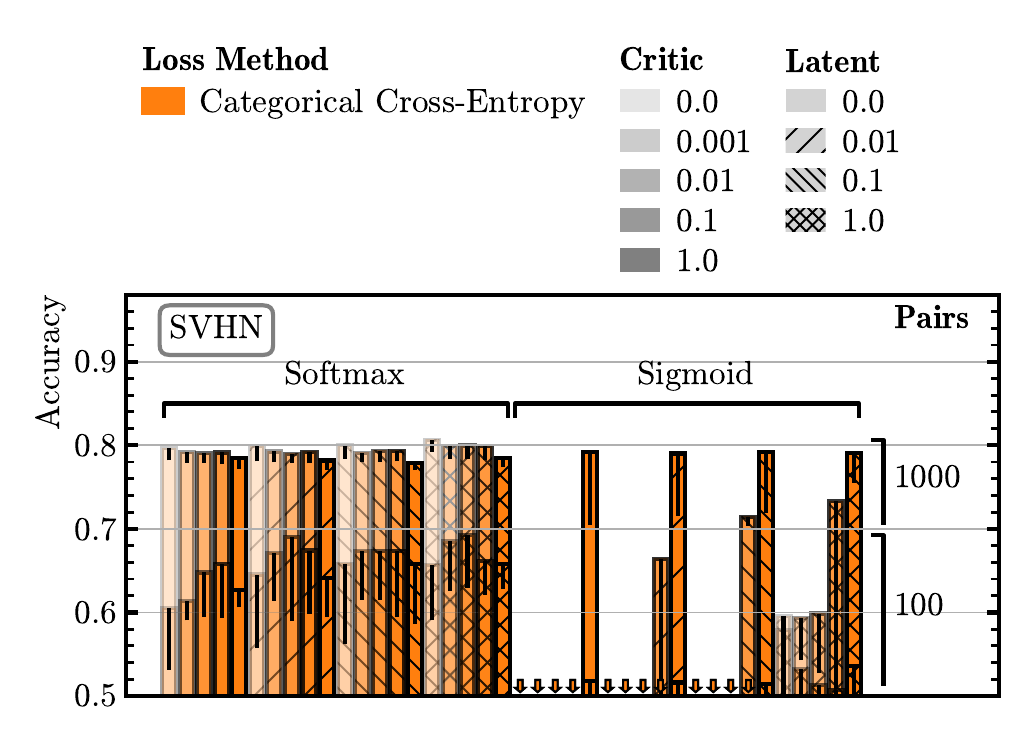} \\

    \includegraphics[scale=\plotscale]{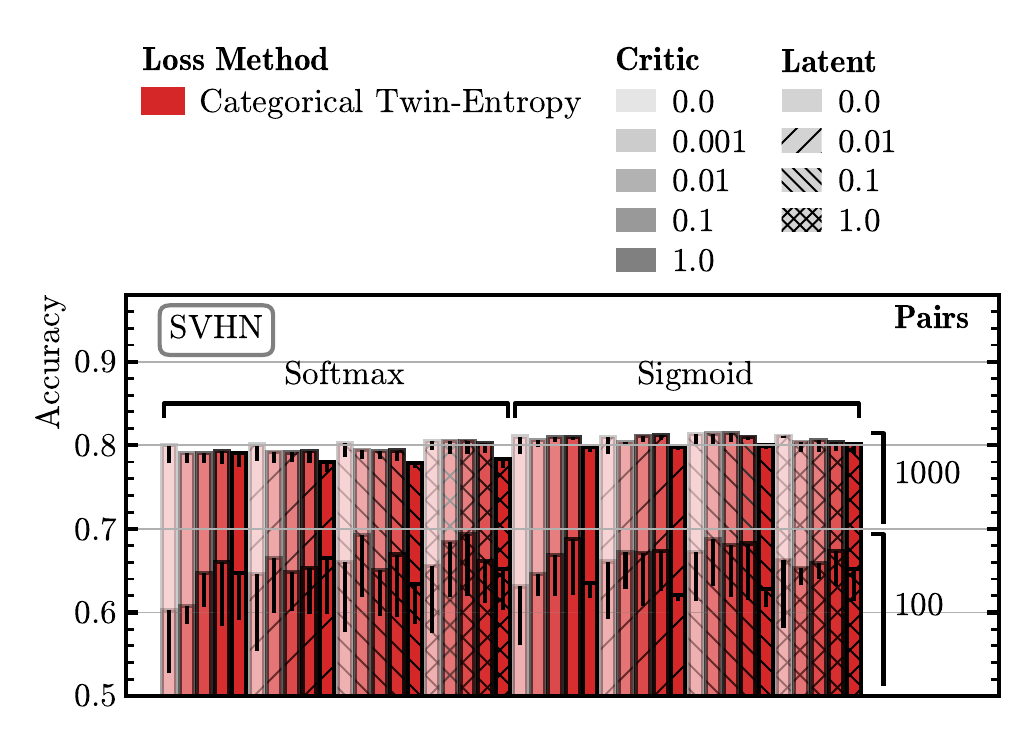} \\

    \includegraphics[scale=\plotscale]{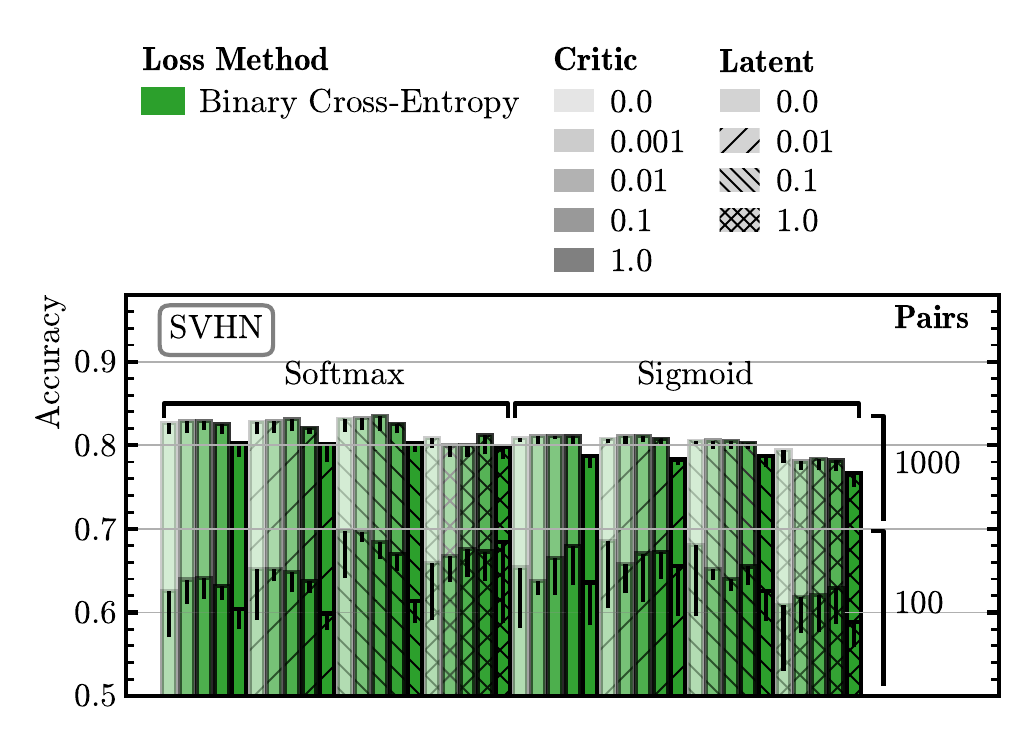} \\
    \caption{All results for the SVHN dataset.}
    \label{fig:svhn_all}
\end{figure}

\begin{figure}[H]
    \centering
    \includegraphics[scale=\plotscale]{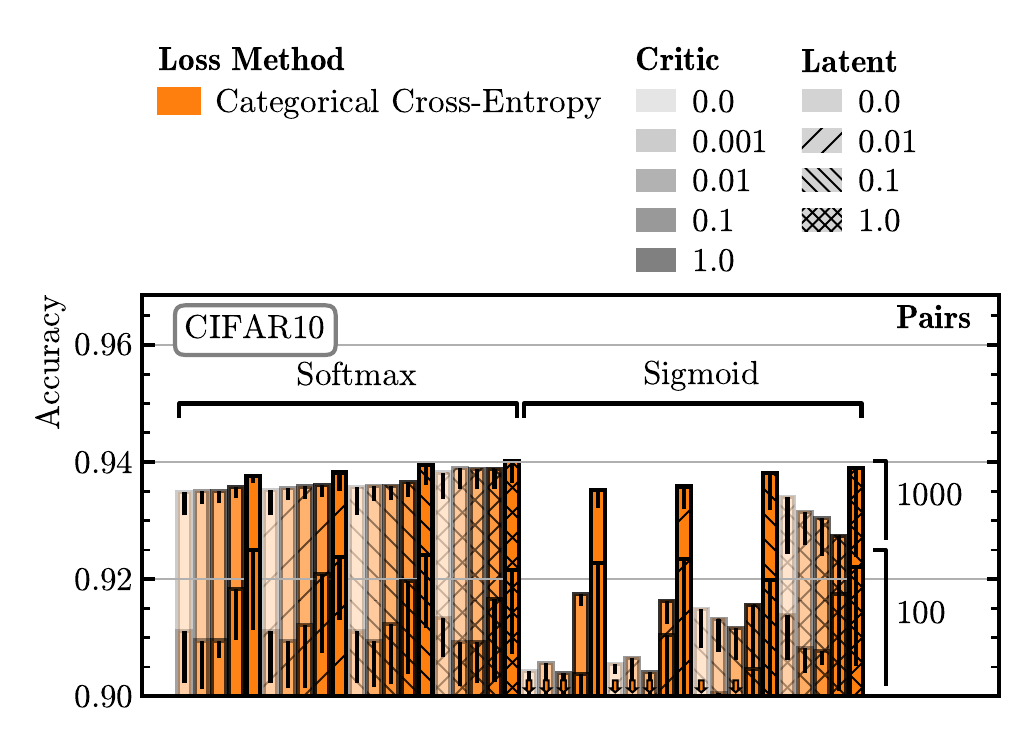} \\

    \includegraphics[scale=\plotscale]{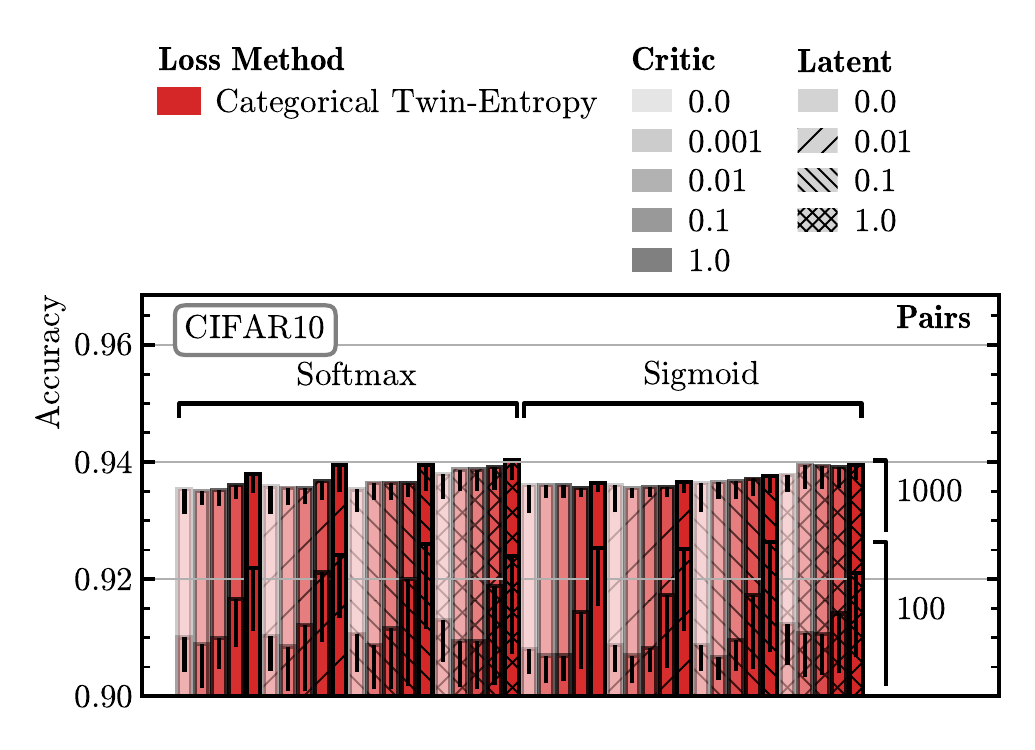} \\

    \includegraphics[scale=\plotscale]{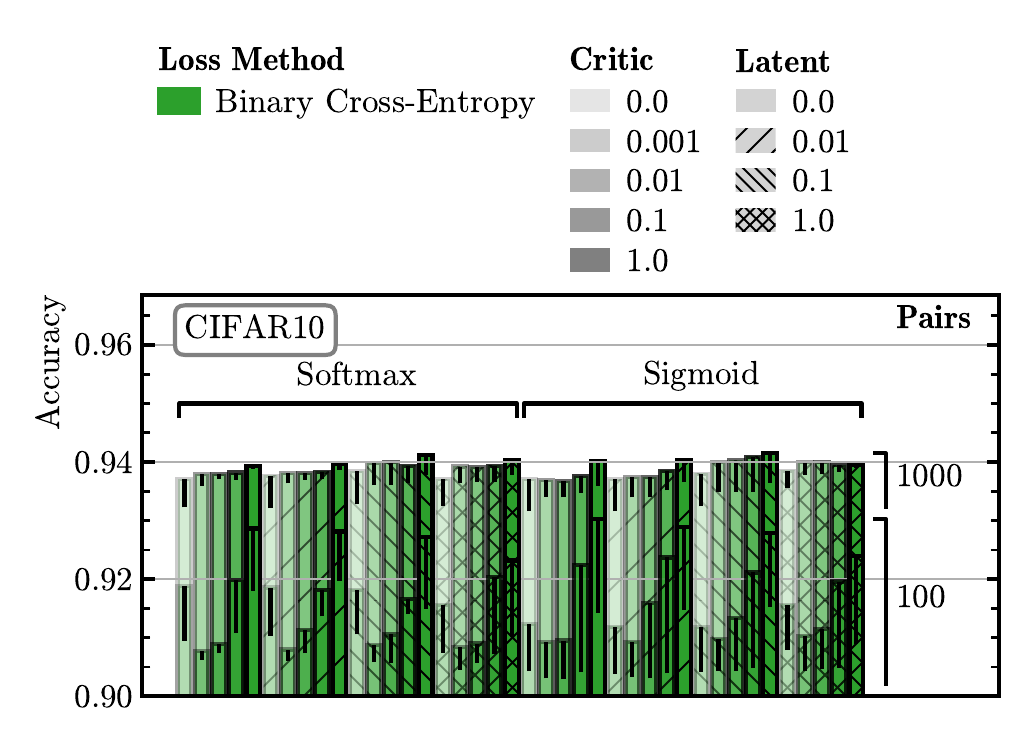} \\
    \caption{All results for the CIFAR10 dataset.}
    \label{fig:cifar10_all}
\end{figure}

\begin{figure}[H]
    \centering
    \includegraphics[scale=\plotscale]{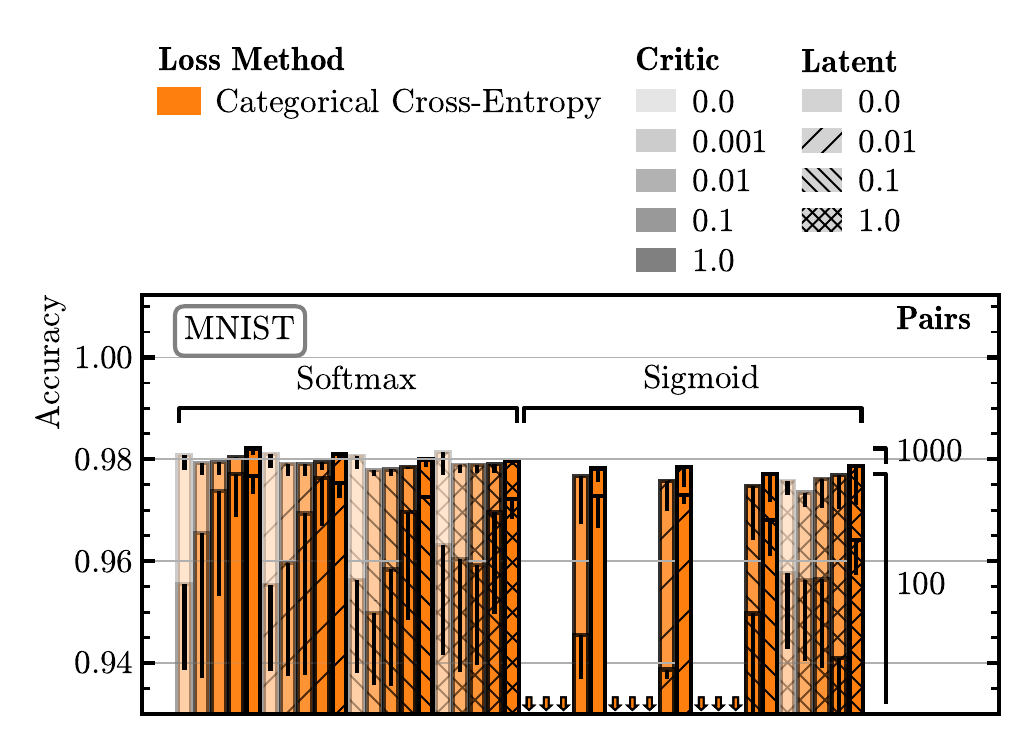} \\

    \includegraphics[scale=\plotscale]{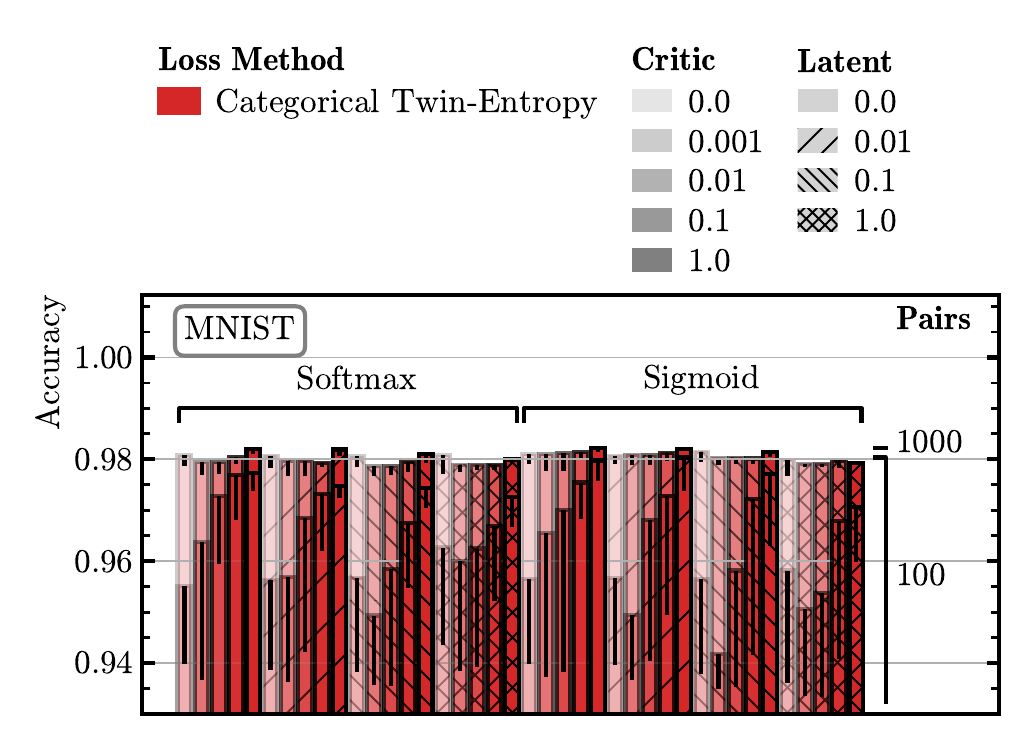} \\

    \includegraphics[scale=\plotscale]{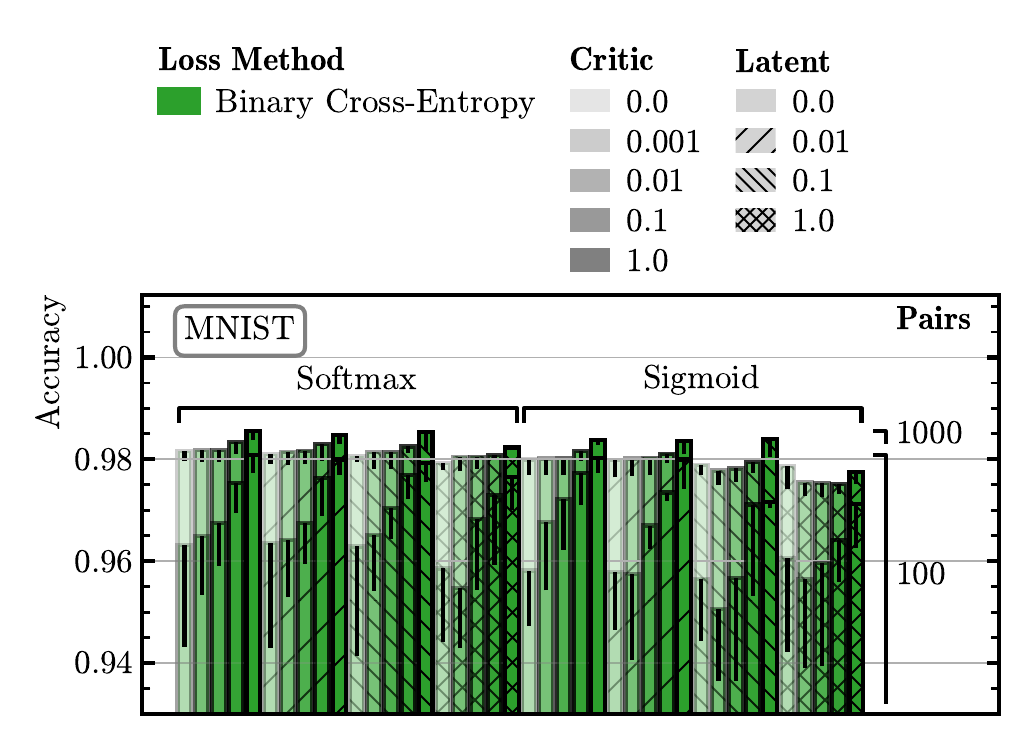} \\
    \caption{All results for the MNIST dataset.}
    \label{fig:mnist_all}
\end{figure}

The results presented in \cref{fig:cifar10_ablation,fig:svhn_all,fig:cifar10_all,fig:mnist_all} show the accuracy of the models trained for every combinations of weights, except for the augmentation weight, which is always set to $\lambda_A = 0$.

\subsection{Ablation plot for CIFAR10}

\begin{figure}[H]
    \centering
    \includesvg[scale=0.42]{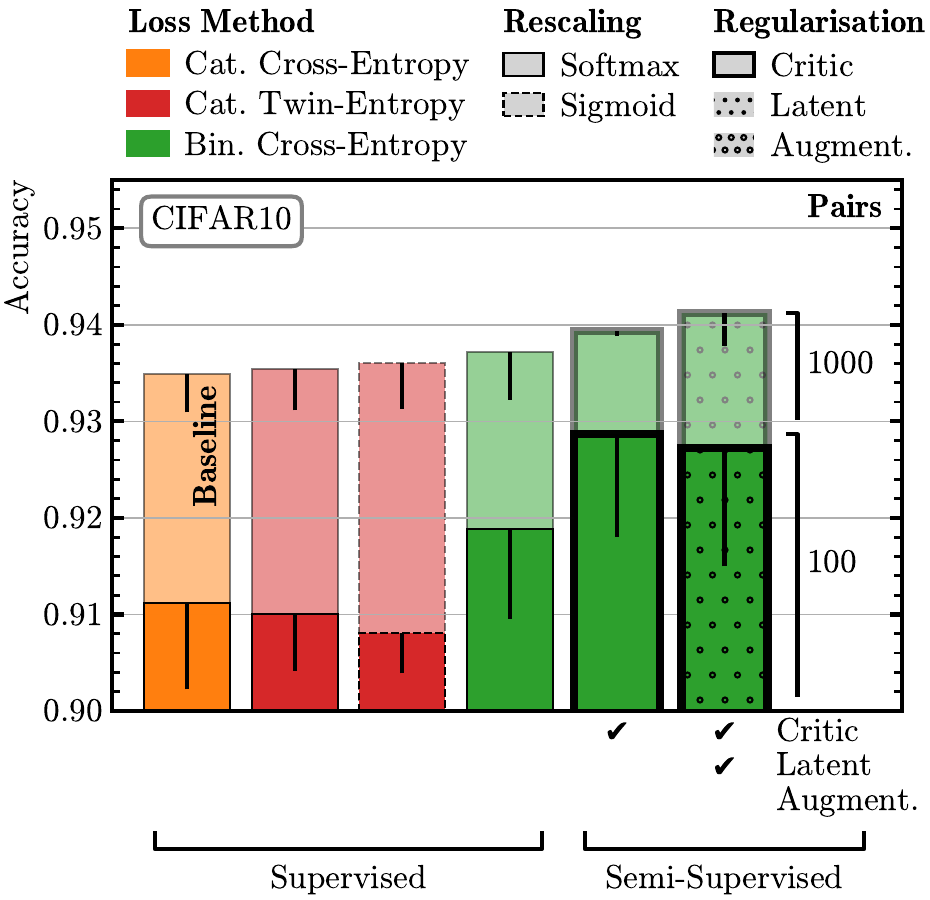}
    \caption{Caption}
    \label{fig:cifar10_ablation}
\end{figure}

\cref{fig:cifar10_ablation} shows the ablation study performed with the CIFAR10 dataset.
The results qualitatively support the claims made in \cref{sec:experiment} and the observations presented in \cref{fig:ablation}.

\subsection{All results}

\end{document}